\documentclass[letterpaper, 10 pt, conference]{ieeeconf}  

\IEEEoverridecommandlockouts                              

\title{\LARGE \bf
Robot Crash Course: Learning Soft and Stylized Falling
}
\usepackage{orcidlink}

\author{
Pascal Strauch\,\orcidlink{0009-0008-5215-3902}, 
David Müller\,\orcidlink{0009-0001-6591-8803}, 
Sammy Christen\,\orcidlink{0000-0002-3511-8565}, 
Agon Serifi\,\orcidlink{0000-0003-4439-0023}, \\
Ruben Grandia\,\orcidlink{0000-0002-8971-6843}, 
Espen Knoop\,\orcidlink{0000-0002-7440-5655}, 
Moritz Bächer\,\orcidlink{0000-0002-1952-1266}
\thanks{$^{1}$Disney Research, Zurich, Switzerland
        {\tt\small {first.last}@disneyresearch.com}}%
}

\usepackage{xcolor}   
\usepackage{graphicx}
\usepackage{siunitx}

\usepackage{multicol}
\usepackage{placeins}
\usepackage{booktabs}
\usepackage{multirow}

\usepackage{amsmath}
\usepackage{amssymb}  
\usepackage{mathtools}
\usepackage{siunitx}
\usepackage{bbm}
\usepackage{bm}
\usepackage{caption}
\newcommand{\vect}{\bm}	

\begin{document}

\twocolumn[{%
\renewcommand\twocolumn[1][]{#1}%
\maketitle
\begin{center}
    \centering
    \captionsetup{type=figure}
    \includegraphics[width=\linewidth]{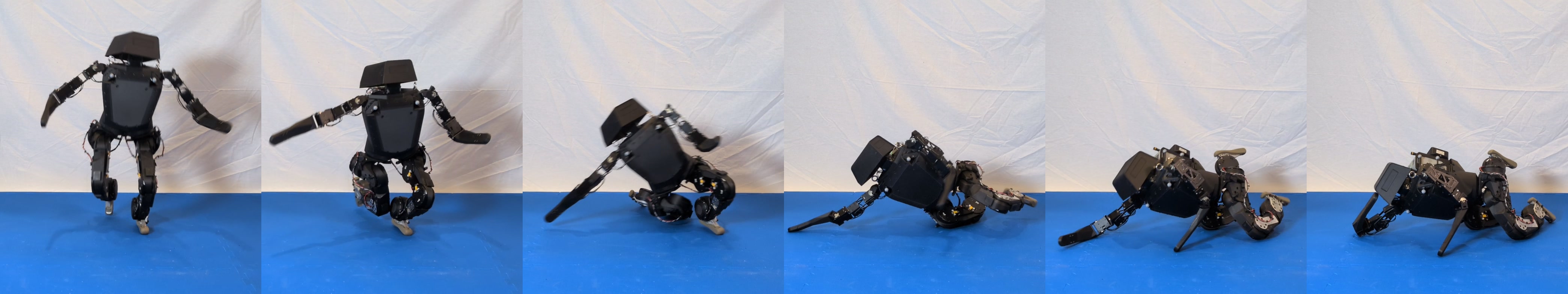}
    \captionof{figure}{We propose a reinforcement learning technique that balances user-guided stylized pose objectives and damage-minimizing soft falling objectives for bipedal and other legged robots.}
    \label{fig:teaser}
\end{center}%
}]

\thispagestyle{empty}
\pagestyle{empty}

\begingroup
\renewcommand{\thefootnote}{}

\footnotetext{All authors are affiliated with Disney Research, Zurich, Switzerland
        {\tt\small {first.last}@disneyresearch.com}}
\endgroup
        
\newcommand{\figref}[1]{Fig.~\ref{#1}}
\newcommand{\secref}[1]{Sec.~\ref{#1}}
\newcommand{\algref}[1]{Algorithm~\ref{#1}}
\newcommand{\eqnref}[1]{Eq.~\eqref{#1}}
\newcommand{\tabref}[1]{Tab.~\ref{#1}}

\newcommand{\mdp}{\mathcal{M}}
\newcommand{\StatesSet}{\mathcal{S}}
\newcommand{\ActionsSet}{\mathcal{A}}
\newcommand{\TransitionSet}{\mathcal{T}}
\newcommand{\GoalSet}{\mathcal{G}}
\newcommand{\goals}{\mathbf{G}}
\newcommand{\discountRate}{\gamma}
\newcommand{\forcevec}{\mathbf{f}}

\begin{abstract}
Despite recent advances in robust locomotion, bipedal robots operating in the real world remain at risk of falling. While most research focuses on preventing such events, we instead concentrate on the phenomenon of falling itself. Specifically, we aim to reduce physical damage to the robot while providing users with control over a robot's end pose. To this end, we propose a robot agnostic reward function that balances the achievement of a desired end pose with impact minimization and the protection of critical robot parts during reinforcement learning. To make the policy robust to a broad range of initial falling conditions and to enable the specification of an arbitrary and unseen end pose at inference time, we introduce a simulation-based sampling strategy of initial and end poses. Through simulated and real-world experiments, our work demonstrates that even bipedal robots can perform controlled, soft falls.
\end{abstract}

\section{INTRODUCTION}
During dynamic motions, legged robots often encounter underactuated contact states that demand continuous dynamic balancing~\cite{wensing2024optimization}. For bipedal robots, this issue is particularly pronounced, as they must control a heavy body on a relatively small area of support. Recent reinforcement learning–based locomotion controllers have made impressive progress in robustness~\cite{zhuang2025humanoid,duan2024learning}, yet the risk of falling in unstructured real-world environments remains substantial. As robots are pushed closer to their limits, much like for humans, certain disturbances or conditions will inevitably cause them to fall. However, unlike humans, robots usually fall in an uncoordinated and uncontrolled manner, leaving delicate components unprotected and breaking the illusion of lifelike motion.

A common approach is to improve controller robustness by adding domain randomization to policy training \cite{grandia2024bdx}, integrating safety-oriented terms into optimization~\cite{romualdi2022online} or reward functions~\cite{radosavovic2024real}, or restricting the uncertainty by reducing the range of capabilities. While such approaches improve stability, they do not guarantee fall prevention in practice and may severely limit a robot's performance or capabilities. Rather than preventing a fall at all costs, we believe it is advantageous to embrace the potential of a fall, providing user control of end poses for stylization and ease of recovery.

In this paper, we therefore explore whether the robot can execute a fall in a controlled and visually appealing manner. Falling is a challenging problem, as it requires performing contact-rich maneuvers within a very short time window and from a wide range of initial states. Moreover, for falling, multiple competing objectives need to be balanced, such as reducing impact, protecting critical components, and achieving desired motion characteristics.

Existing research on robot falling mostly focuses on a single objective or a controlled scenario. Once a failure is detected, common strategies are to \textit{freeze} the actuators with high gains~\cite{atkeson2015nofalls} or achieve a compliant reaction using low gains. However, both approaches offer limited controllability over the resulting motion and suffer from high impact. More involved solutions often rely on hand-crafted fall strategies, such as executing predefined falling motions \cite{Ukemi_Fujiwara2002} or tracking predefined contact sequences \cite{UkemiWalk_Ogata2007}. This idea has recently been broadened to adaptive contact sequences, but remains restricted to a single falling direction \cite{AdaptiveContact_Ha2015, LearningUnified_Kumar2017} or requires manual considerations tailored to specific fall scenarios, such as falling forwards or backwards \cite{ishida2004analysis}.

In contrast, our method not only reduces overall impact forces but also provides fine-grained user control, through the specification of critical components to protect and desired end poses for the robot to reach. This can be used for artistic control, as shown here, but could also serve as a starting pose for a recovery policy. We propose a reinforcement learning (RL) solution that offers adjustable trade-offs between damage reduction and pose objectives. To generalize across a wide range of user-specified end poses, we propose a physics-informed sampling strategy that comprehensively covers the distribution of initial and final states. Importantly, by leveraging reinforcement learning, our approach supports a wide variety of falling scenarios. 

In our experiments, we compare our method quantitatively with standard falling strategies, showing that our approach results in softer falls. Through an ablation study in simulation, we demonstrate how our proposed method leads to controlled falling while adhering to landing in desired poses, with a user-defined trade-off between the two. In real-world experiments, we qualitatively demonstrate that our policies lead to falls without damage. 
To the best of our knowledge, this is the first general approach that demonstrates user-controlled falling of a bipedal robot in the real world. While we focus our evaluation on bipedal systems due to their inherent instability, our modeling is agnostic to the number of legs. 

Succinctly, we contribute: 
\begin{itemize}
    \item A learning-based technique that balances impact minimization with a user-defined end pose, providing artistic control over a fall and facilitating a successful recovery. 
    \item A sampling strategy of initial and end poses, enabling the training of a general falling policy that allows a user to specify an unseen and desired end pose at inference. 
    \item Extensive ablations of our method in simulation and qualitative evaluation on a bipedal robot, highlighting the utility of our approach. 
\end{itemize}

\section{RELATED WORK}

\subsection{Soft Falling}

Initial works on bipedal falling rely on hand-crafted strategies and predefined motions. A common approach is to treat falling as a controlled event that is managed with a sequence of carefully designed actions. For example, controllers can trigger specific joint trajectories, such as bending the robot's knees and extending its arms to reduce impact forces~\cite{ishida2004analysis,wang2017real}, or tracking UKEMI-inspired falling motions~\cite{Ukemi_Fujiwara2002,UkemiWalk_Ogata2007}. Alternatively, the robot can be guided by a predefined contact sequence~\cite{TripodFall_Yun2014}. As a complementary strategy, the gains of the actuators can be softened, making the joints more compliant to passively absorb impacts~\cite{ishida2004analysis,samy2015falls}.

A main limitation of these earlier works is their focus on relatively slow, locomoting robots and falls occurring primarily in the sagittal plane. As robots become capable of more dynamic motions~\cite{serifi2024vmp, liao2025beyondmimic}, the likelihood of multi-directional falls with high impact forces increases.

Recent advancements in RL allow for more general, flexible, and robust methods that require fewer assumptions. 
By allowing for adaptive and learned contact sequences, various individual fall strategies can be unified \cite{AdaptiveContact_Ha2015,LearningUnified_Kumar2017}, and scale beyond the simplification of sagittal plane falls \cite{QP_Samy2017}. There has been more recent focus on quadruped falling policies~\cite{wang2023guardians,AlmaFall_Ma2023}.
ALMA~\cite{AlmaFall_Ma2023}, for example, provides a general framework that assigns time-varying damage-reduction rewards across the different phases of a fall. 

We extend upon the related works by leveraging the strength of RL and propose a general learning framework for soft falling that covers diverse falling scenarios. Our method accounts for sensitive robot parts, enabling the policy to minimize impacts on critical components, without manually specifying falling motions or contact sequences.

\subsection{Stylized Falling}

For applications in human-robot interaction or the entertainment industry, lifelike and stylistic robot motions become important~\cite{grandia2024bdx,christen2025autonomous,alvarez2025learning}. In character animation, motion is typically defined by keyframes, with intermediate poses generated using various interpolation methods. Those methods range from simple parametric curves \cite{lee1999hierarchical} to sophisticated learning-based interpolation techniques \cite{agrawal2024skel}. Keyframes were also explored in RL to sparsely define a robot's motion~\cite{zargarbashi2024robotkeyframing}. However, so far, artistic keyframes have only been applied in controlled settings, and not in the context of a falling objective. In fact, most works in simulated character control aim to prevent falling at all costs, either by introducing non-physical fictitious forces~\cite{yuan2020residual}, employing early termination techniques~\cite{peng2018deepmimic}, or explicitly penalizing falling through a reward~\cite{serifi2024vmp}.

We extend the capabilities of a falling policy by combining soft falling with style guidance, an aspect particularly relevant for human-oriented applications. Specifically, our framework enables steering the fall towards stylistic end poses while reducing the impact forces caused by it.

\section{OVERVIEW}

Falling motions are challenging as they arise from diverse and unstable initial states. Our goal is to enable a robot to perform a controlled fall across such conditions, while minimizing impact and reaching a user-specified, stylized end pose. Specifically, a user specifies two inputs: the relative sensitivity between robot components that are fixed at training time, and a desired end pose that is specified at inference time and should be reached by the robot once at rest, irrespective of the initial falling condition. For example, even when stumbling backwards, the robot should be able to achieve an end pose where it lies on the front side of the pelvis while protecting its head with its arms. 

End poses can serve different goals. For instance, artists can define expressive poses to enable falls with an intended stylistic effect, turning a perceived failure into a believable motion. Moreover, end poses can be chosen to serve as suitable starting poses for recovery policies, facilitating a seamless transition into a standing pose \cite{tao2022getup, he2025learning}.

Note that this work focuses exclusively on a graceful falling behavior of a robot; not on deciding whether a fall should occur.

\section{METHOD}

\begin{figure}[t]
    \centering
        \includegraphics[width=\linewidth]{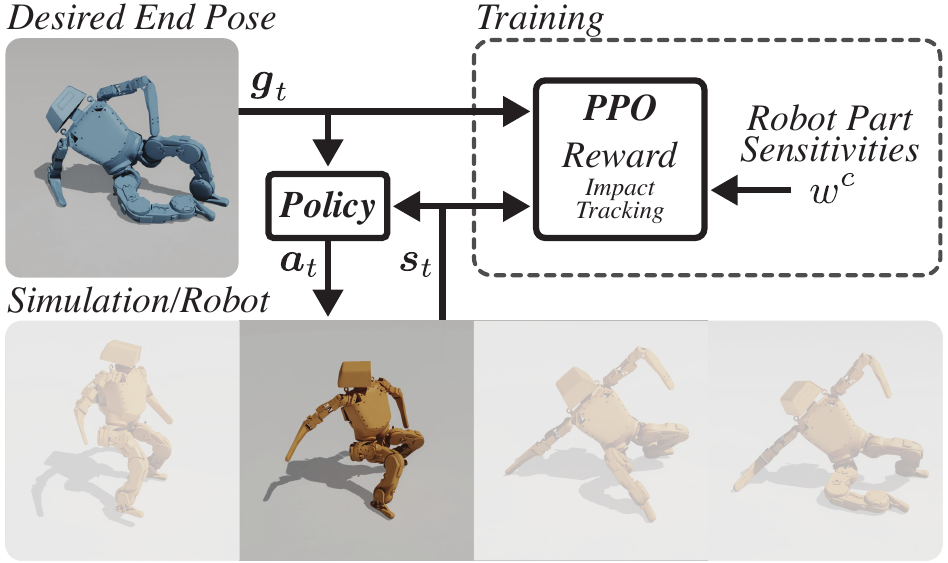}
    \caption{\textbf{Method Overview.} We leverage reinforcement learning to train a robust falling policy (right). Our method learns to balance \textbf{impact minimization} with reaching a \textbf{desired end pose} through our reward formulation, which considers user-specified robot part sensitivities. During inference (left), the policy is guided by a user-specified end pose, while simultaneously minimizing impact.}
    \label{fig:overview}
\end{figure}

To achieve our goals as outlined above, our method trains a policy via reinforcement learning, relying on a reward function that balances impact minimization with pose objectives (refer to Fig.~\ref{fig:overview}). Before describing our rewards, sampling strategy, and initialization procedure, we introduce the domain-specific states, goals, and actions.

\subsection{Reinforcement Learning Setup}
\label{sec:state}

Our goal is to determine a sequence of actions $\vect{a}_t$ through a policy $\pi(\vect{a}_t \vert \vect{s}_t, \vect{g}_t)$ that transitions the robot from an initial state $\vect{s}_0$ into a final state $\vect{s}_{T}$, taking on a user-specified end pose, with $t \in [0, T]$. The actions $\vect{a}_t$ are joint position setpoints for proportional-derivative (PD) controllers, and the proprioceptive state is
\begin{equation}
    \vect{s}_t := \big({\vect{\theta}}_t, {\vect{v}}_t, {\vect{\omega}}_t, \vect{q}_t, \dot{\vect{q}}_t, \vect{a}_{t-1}, \vect{a}_{t-2}\big), 
\end{equation}
where ${\vect{\theta}}_t$ is the root orientation, represented with a unit quaternion, ${\vect{v}}_t$ and ${\vect{\omega}}_t$ are the root's linear and angular velocities, and $\vect{q}_t$ and $\dot{\vect{q}}_t$ are its joint angles and joint angular velocities. 

\noindent The time-varying goal 
\begin{equation}
    \vect{g}_t := \big(\hat{\vect{\theta}}_t,\hat{\vect{q}}_t\big)
\end{equation}
is derived from the user-specified end pose $\vect{g} = \big(\hat{\vect{\theta}},\hat{\vect{q}}\big)$ as outlined below, where $\hat{\vect{\theta}}_t$ is the robot's target root orientation and $\hat{\vect{q}}_t$ its target joint configuration.

To make our policy invariant under the robot's global pose, we represent the root orientation in states and goals in the local path frame, and the root's linear and angular velocities w.r.t.~the root frame. The path frame is defined with the root at the origin, x- and y-axes in the horizontal plane, and the x-axis aligned with the root’s facing direction.

Note that we do not vary the end goal but apply the time-varying transformation from the end pose to path-relative coordinates. This ensures that the policy has sufficient freedom to pursue competing objectives and to reach the desired final state while minimizing impacts.

During training, we sample initial states $\vect{s}_0$ and end poses $\vect{g}$ as described in Secs.~\ref{sec:procedural} and ~\ref{sec:initialization}.

\subsection{Reward Design}
\label{sec:reward}
The reward function is designed to balance accurate \emph{end pose tracking} with \emph{soft impact}, supplemented by \emph{regularization} and a constant \emph{positive offset}
\begin{equation}
r_t = r^{\text{tracking}}_t + r^{\text{impact}}_t + r^{\text{regularization}}_t + r^{\text{offset}}.
\label{eq:reward}
\end{equation}

A detailed breakdown of the weighted reward terms is provided in Tab.~\ref{tab:rewards}, where hats $\hat{\cdot}$ denote target quantities derived from $\vect{g}$.

\begin{table}[tb]
\begin{center}
    \caption{\textbf{Weighted Reward Terms.} To penalize contact forces, we sum up all contact forces that act on a component $c$ in the force vectors $\vect{f}^c_t$, then multiply them with the component's sensitivity weight $w^c$. $\dot{\vect{v}}_t$ is the linear acceleration of the root, and $\vect{\tau}_t$ and $\ddot{\vect{q}}_t$ are joint torques and accelerations. For end pose tracking of the root orientation, we apply Rodrigues' rotation formula to convert unit quaternions to rotation matrices, measuring differences in yaw rotations between the simulated and the goal state ($\vect{e}_z$ is the unit vector along the z-axis).}
    \label{tab:rewards}
    \begin{tabular}{l | l | l}\toprule 
    \textbf{Name}                  &  \textbf{Reward Term} & \textbf{Weight}\\
    \midrule
    \midrule
    \multicolumn{3}{c}{\textit{Impact}}  \\
    \midrule
    Contact forces & $-\sum_{\text{comp. } c} \lVert w^c \vect{f}^c_t \rVert_{\infty}^2$ & $200$ \\
    Root acc. &  $-\lVert \dot{\vect{v}}_t \rVert_2 ^2 $ & $0.2$ \\
    \midrule
    \multicolumn{3}{c}{\textit{End Pose Tracking}}  \\
    \midrule
    Root orientation & $-u(t)\lVert {\vect{R}}(\vect{\theta}_t)^{T}\vect{e}_z - {\vect{R}}(\hat{\vect{\theta}_t})^{T}\vect{e}_z \rVert_2^{2}$ & $20.0$ \\
    Joint positions &  $-u(t)\lVert \vect{q}_t - \hat{\vect{q}}_t \rVert_2 ^2 $ & $ 1.0 $ \\
    \midrule
    \multicolumn{3}{c}{\textit{Regularization}}  \\
    \midrule
    Joint torques                 &  $-\lVert \vect{\tau}_t \rVert_2 ^2 $ & $1.0 \cdot 10^{-3}$ \\
    Joint acc.           &  $-\lVert \ddot{\vect{q}}_t \rVert_2 ^2 $ & $7.5 \cdot 10^{-7}$ \\
    Action rate        &  $-\lVert \vect{a}_t - \vect{a}_{t-1} \rVert_2 ^2 $ & $ 0.1 $ \\
    Action acc.      &  $-\lVert \vect{a}_t - 2\vect{a}_{t-1} + \vect{a}_{t-2} \rVert_2 ^2 $ & $ 0.05 $ \\
    \midrule
    \multicolumn{3}{c}{\textit{Positive Offset}}  \\
    \midrule
    Positive offset  &  $1.0$ & $50$ \\
    \bottomrule
    \end{tabular}
\end{center}
\end{table}

To promote soft impacts, we adopt an impact reward (\tabref{tab:rewards}, \emph{top}) inspired by prior work on quadrupedal falling~\cite{AlmaFall_Ma2023}. We extend the contact force reward by scaling contact forces of robot components with non-negative sensitivity weights $w^{c}$. Root acceleration is also penalized to discourage abrupt motion, regardless of contact.

The tracking reward $r^{\text{tracking}}_t$ (\tabref{tab:rewards}, \emph{middle}) compares the simulated robot pose to a target end pose and incentivizes the policy to reach this pose.  Specifically, the tracking reward combines a joint tracking and a global yaw-invariant orientation term. To encourage the policy to initially focus on reducing impact forces, before smoothly transitioning to pose tracking, the tracking reward is modulated with a time-dependent cubic spline $u(t)$, which interpolates the reward between $0.0$ and $1.0$ over the blending duration $T_{\text{blend}}$:
{
\medmuskip=0mu
\thinmuskip=1mu
\thickmuskip=0mu
\delimitershortfall=-1pt
\begin{equation}
u(t) =
\begin{cases}
-2\left(\dfrac{t}{T_{\text{blend}}}\right)^3 + 3\left(\dfrac{t}{T_{\text{blend}}}\right)^2, & 0 \leq t \leq T_{\text{blend}}, \\[1.0ex]
1.0, & t > T_{\text{blend}}.
\end{cases}
\end{equation}
}
The parameter $T_{\text{blend}}$ is empirically determined to balance impact reduction and tracking performance, and satisfies $T_{\text{blend}} \leq T$. In practice, we find that extending learning beyond the interpolation time helps the policy come to rest and maintain the final pose without jitter. 

Following prior work~\cite{grandia2024bdx, peng2018deepmimic} for imitation learning, we add regularization rewards (\tabref{tab:rewards}, \emph{bottom}) to penalize excessive joint torques and encourage smooth actions, helping to avoid vibrations and unnecessary effort. Finally, a constant positive reward ensures that the agent observes positive rewards from the start of training, which facilitates learning~\cite{sullivan2023rewardscalerobustnessproximal}.

\subsection{Sampling-Based End Pose Generation}
\label{sec:procedural}
To enable user control over a wide range of end poses at inference time, we introduce a physics-informed sampling mechanism to generate a dataset of statically stable and feasible robot configurations.

We begin by sampling random joint configurations within feasible limits and discarding all configurations that result in self-collision between robot components. Next, we sample the robot's root orientation by applying first pitch and then yaw rotations around the robot's axis, both over the full range of $\pm \ang{180}$.

To obtain statically stable end poses, each sampled configuration is initialized and dropped from a predefined height of \SI{0.04}{\meter} with actuators frozen (high gains with fixed setpoint), until the robot comes to rest, as visualized in the supplemental video material.

This procedure, however, can produce a biased distribution: certain root orientations, such as poses on the robot’s back, may be overrepresented, while others, like poses on the side, are underrepresented. To mitigate this bias, we iteratively sample new poses while discarding those that are already sufficiently represented, ensuring uniform coverage across root orientation bins.

We leverage Isaac Sim~\cite{NVIDIA_Isaac_Sim} to perform collision detection and to let frozen robots settle into statically-stable configurations, enabling GPU-accelerated generation of large batches of physically-feasible end poses.

\begin{table}[t!]
\centering
\caption{\textbf{Initial Robot State Ranges.} During training, we cover a wide range of initial falling conditions by sampling from the ranges below.}
\label{tab:initial_state}
\begin{tabular}{l l}
\toprule
\textbf{Variable} & \textbf{Range} \\
\midrule
\midrule
Roll, Pitch  & $[-30, 30]$~\si{degrees}\\
Linear root velocity  & $[-2.0, 2.0]~\si{\meter\per\second}$ \\
Angular root velocity & $[-0.5, 0.5]~\si{\radian\per\second}$ \\
Joint velocities      & $[-0.5, 0.5]~\si{\radian\per\second}$ \\
\bottomrule
\end{tabular}
\end{table}

\begin{figure*}[t!]
 \centering
  \begin{minipage}{1\linewidth}
  \centering
  \begin{minipage}{0.19\linewidth}
   \centering
   \includegraphics[width=\linewidth]{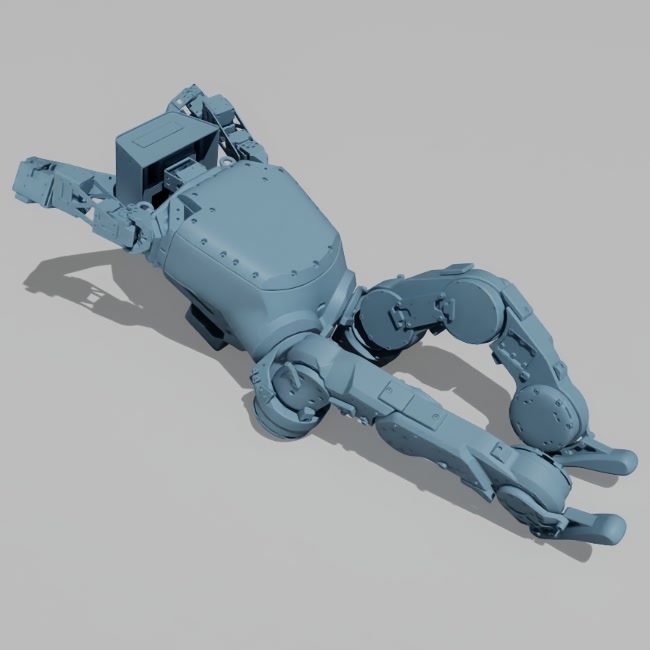}
   \\ 
  \end{minipage}
  \begin{minipage}{0.19\linewidth}
   \centering
   \includegraphics[width=\linewidth]{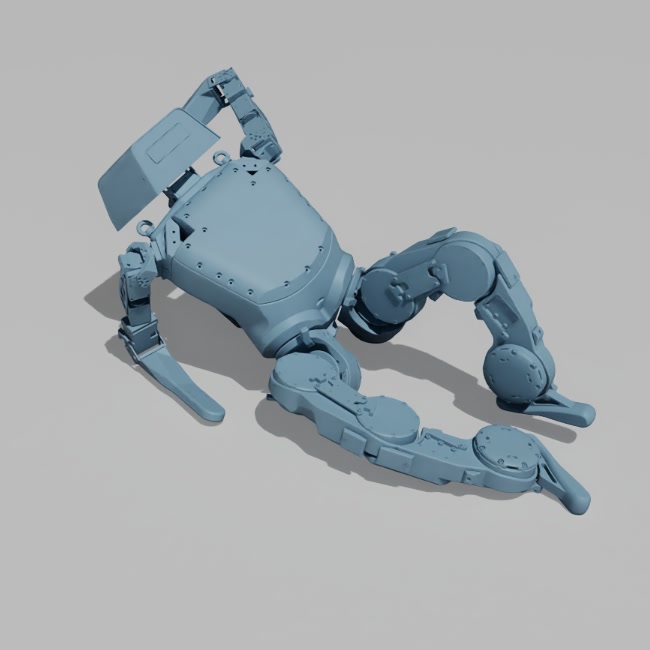}
   \\ 
  \end{minipage}
  \begin{minipage}{0.19\linewidth}
   \centering
   \includegraphics[width=\linewidth]{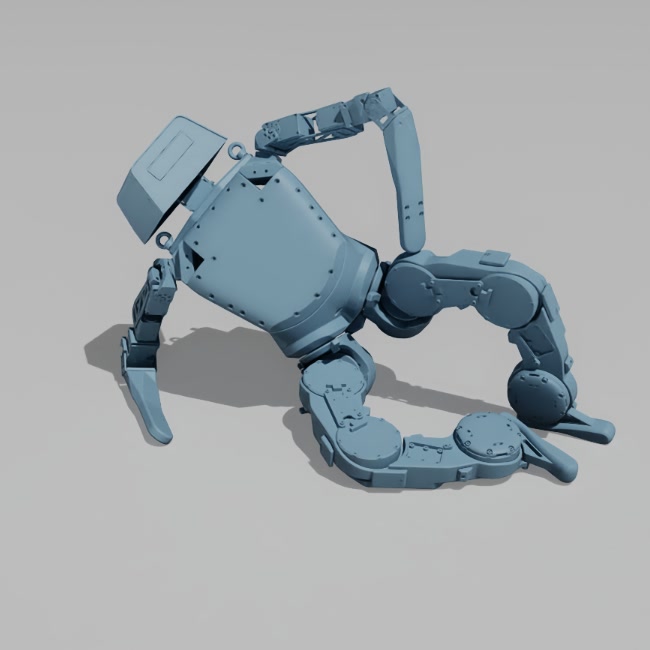}
   \\ 
  \end{minipage}
  \begin{minipage}{0.19\linewidth}
   \centering
   \includegraphics[width=\linewidth]{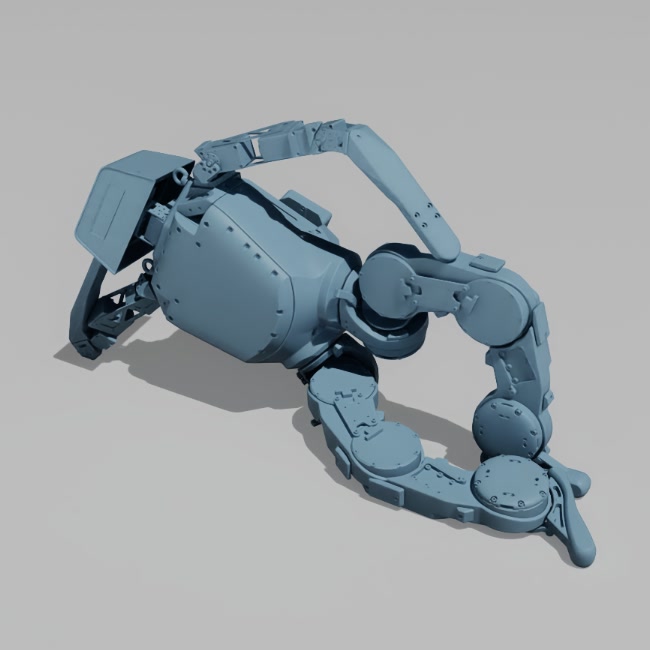}
   \\ 
  \end{minipage}
  \begin{minipage}{0.19\linewidth}
   \centering
   \includegraphics[width=\linewidth]{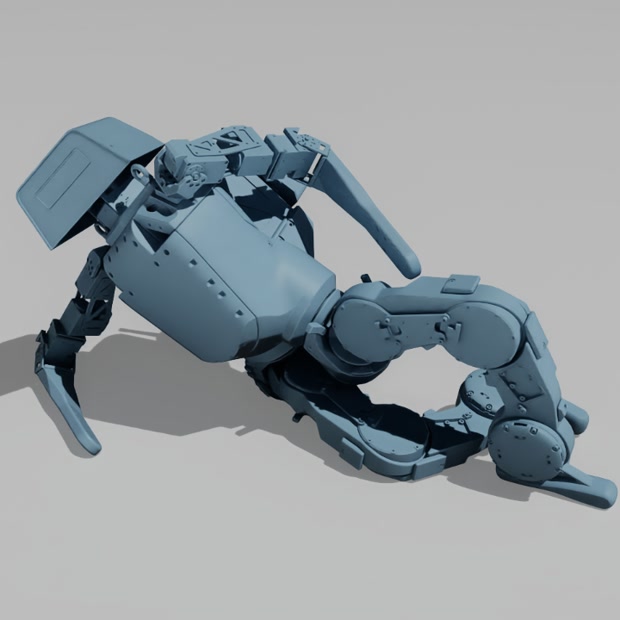}
   \\ 
  \end{minipage}
   \\~\\ \vspace{1mm}
  \begin{minipage}{0.19\linewidth}
   \centering
   \includegraphics[width=\linewidth]{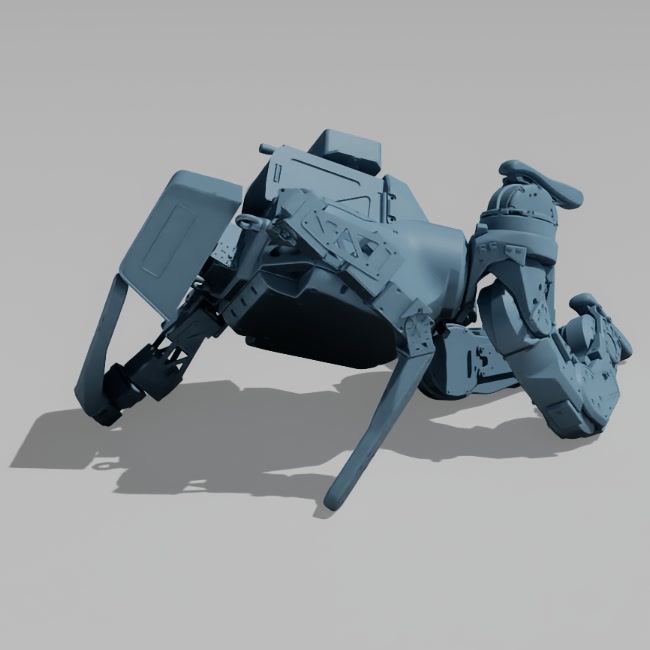}
   \\ 
  \end{minipage}
  \begin{minipage}{0.19\linewidth}
   \centering
   \includegraphics[width=\linewidth]{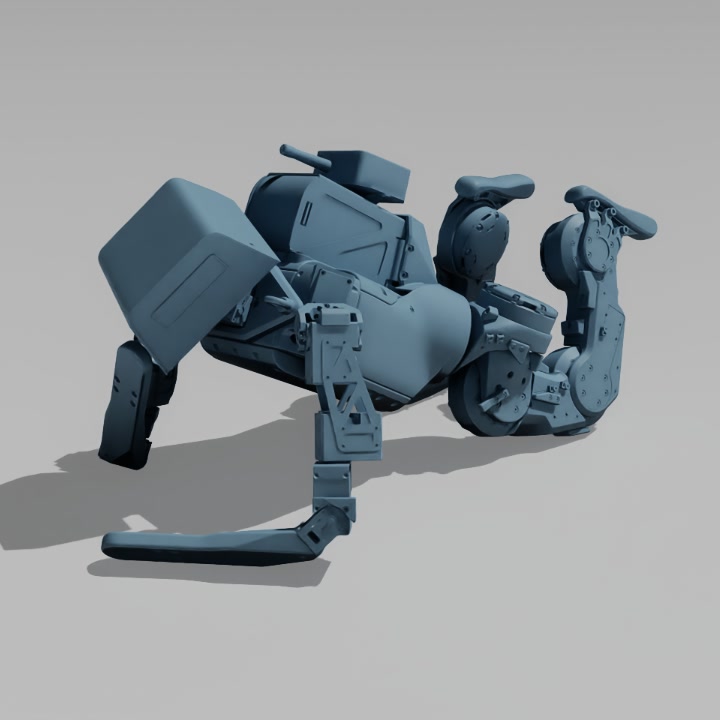}
   \\ 
  \end{minipage}
  \begin{minipage}{0.19\linewidth}
   \centering
   \includegraphics[width=\linewidth]{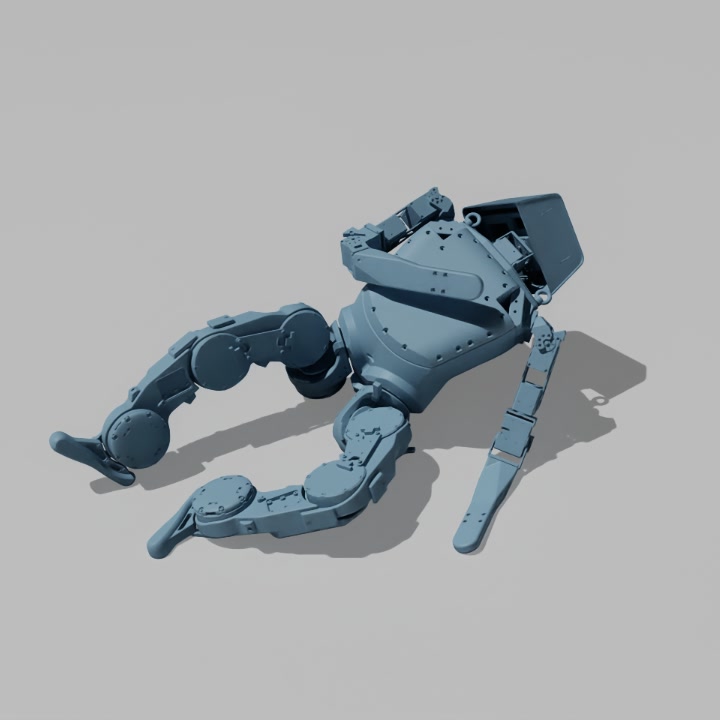}
   \\ 
  \end{minipage}
\begin{minipage}{0.19\linewidth}
   \centering
   \includegraphics[width=\linewidth]{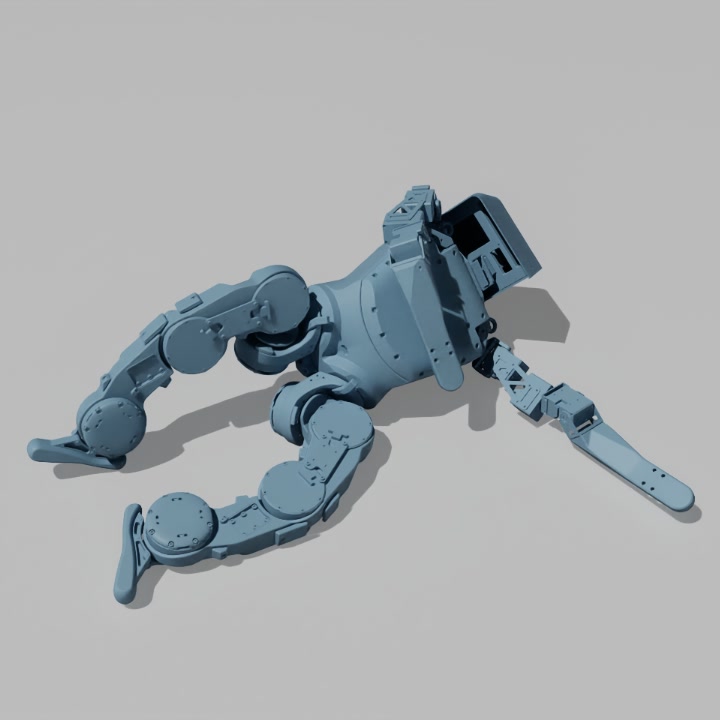}
   \\ 
  \end{minipage}
  \begin{minipage}{0.19\linewidth}
   \centering
   \includegraphics[width=\linewidth]{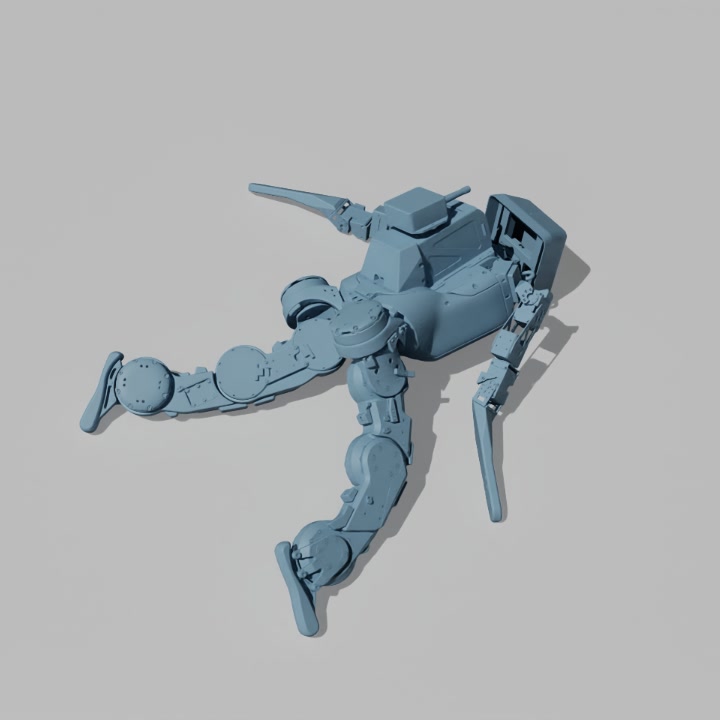}
   \\ 
  \end{minipage}
  
 \end{minipage}
 \caption{\textbf{Artist-Designed End Poses.} Visual examples of the 10 artist-designed end poses used in our experiments.}
 \label{fig:artistic}
\end{figure*}

\subsection{Robot Initialization}
\label{sec:initialization}

To cover a wide range of possible initial falling states in which the policy may be activated, we randomize the initial conditions at the beginning of each episode. We sample a root orientation together with a joint configuration within feasible limits, and filter all poses to avoid ground penetration or self-collisions. Since our method is invariant to the global yaw angle, only the pitch and roll are varied, sampled in pitch–roll order. To further increase variability and to mimic the effects of external perturbations and unstable (i.e., falling) starting conditions, we also assign initial velocities to both the joints and the root. The corresponding parameter ranges are summarized in \tabref{tab:initial_state}.

\begin{table}[tb]
\begin{center}
\caption{\textbf{PPO hyperparameters.} The hyperparameters used to train the falling policy.}
\footnotesize
\begin{tabular}{l|l}
\toprule
\textbf{Param.}      & \textbf{Value}     \\ \midrule \midrule
Num. iterations & $75\,000$ \\
Batch size
$(\text{num.     envs.}\times\text{steps})$ & $4096\times24$  \\
Num. mini-batches & $4$  \\  
Num. epochs  &  $5$  \\
Clip range & $0.2$  \\
Entropy coefficient & $0.0$ \\
Discount factor  & $0.99$  \\
GAE discount factor & $0.95$  \\
Desired KL-divergence & $0.01$  \\
Max gradient norm & $1.0$  \\
\bottomrule
\end{tabular}
\label{tab:hyperparams_rl}
\end{center}
\end{table}
\begin{table}[t!]
\centering
\caption{\textbf{Disturbance Forces.} We add random forces and torques to each specified body part, drawn from uniform distributions with the magnitudes listed below and applied per dimension. Disturbances are applied for a random ``on'' duration, followed by a random ``off'' duration before the next application.}

\begin{tabular}{ll|ll}
\toprule
\textbf{Param.}  &  & \textbf{Range}  \\ \midrule \midrule
Body                          &     & Hips, Feet, Elbows & Pelvis, Head \\
Force [\SI{}{\newton}]   & XY  &  [0.0, 5.0]  & [0.0, 5.0] \\
                              & Z   &  [0.0, 5.0]  & [0.0, 5.0] \\
Torque [\SI{}{\newton\meter}]  & XY  &  [0.0, 0.25] & [0.0, 0.25] \\
                              & Z   &  [0.0, 0.25] & [0.0, 0.25] \\
Duration [\SI{}{\second}] & On  &  [0.25, 2.0] & [2.0, 10.0] \\
                              & Off &  [1.0, 3.0]  & [1.0, 3.0] \\ \bottomrule
\end{tabular}
\label{tab:dist_forces}
\end{table}

\begin{figure*}[t!]
 \centering
  \begin{minipage}{0.92\linewidth}
  \centering
      \begin{minipage}{0.49\linewidth}
       \centering
       \includegraphics[width=\linewidth]{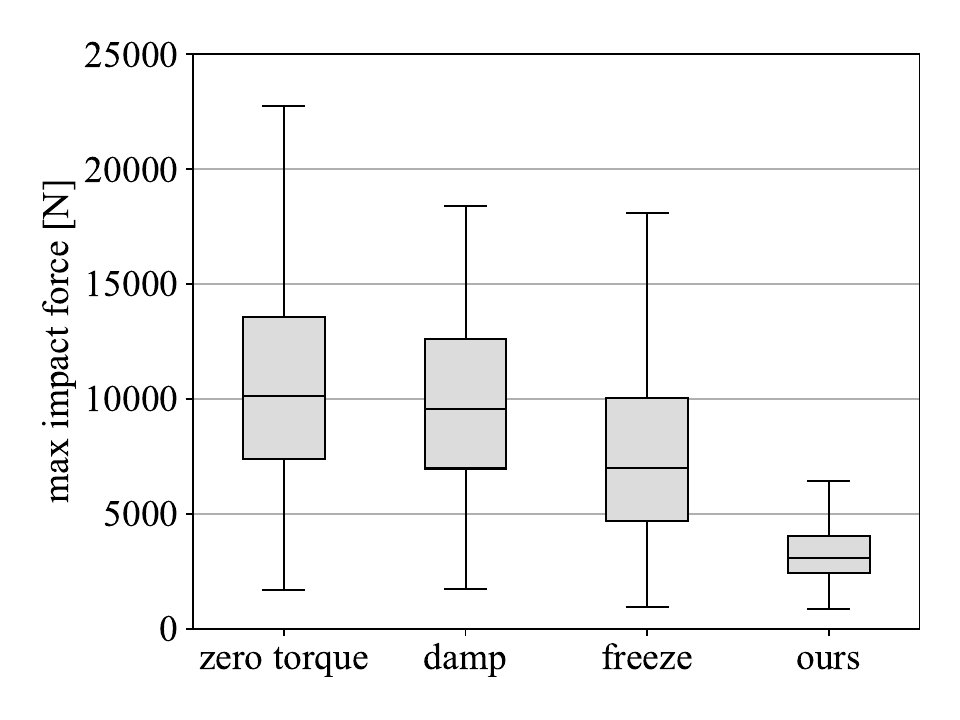}
       \\ 
      \end{minipage}
      \begin{minipage}{0.49\linewidth}
       \centering
       \includegraphics[width=\linewidth]{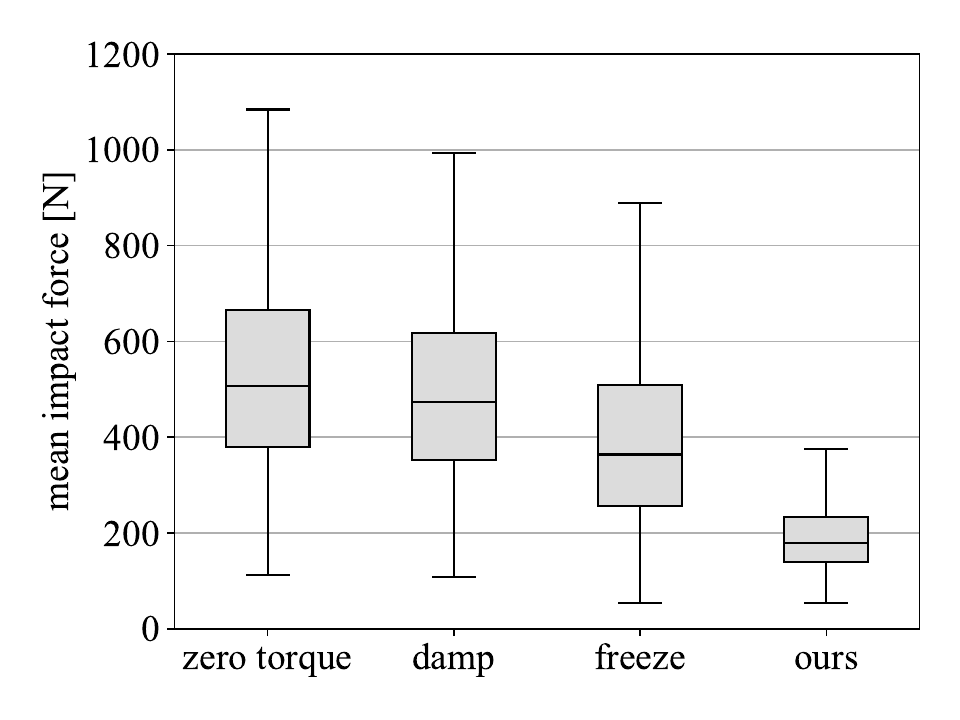}
       \\ 
      \end{minipage}
 \end{minipage}
  \caption{\textbf{Impact Analysis.} Comparison of maximal (left) and mean (right) impact forces across body parts between standard falling strategies and our method.}
 \label{fig:impact_force}
\end{figure*}

\section{EXPERIMENTS AND RESULTS}

We first outline implementation (\secref{sec:implementation_details}) and experimental details (\secref{sec:experimental_details}). Our evaluations are then organized into four parts. First, we compare our method against default falling strategies (\secref{sec:comparison_failsafe}). Next, we perform extensive ablations of our approach that highlight the effectiveness of our reward formulation and the sampling-based end pose generation (\secref{sec:ablations}). We then showcase how the robot part sensitivity weights affect the resulting impact forces (\secref{sec:critical_comp}). Finally, we demonstrate the transferability of our approach from simulation to the real world through experiments with a bipedal robot (\secref{sec:sim2real}).

\subsection{Implementation Details}
\label{sec:implementation_details}
In the following, we provide details on the network architecture, training durations, and other relevant information to facilitate reproducibility. 

We train the falling policy using PPO~\cite{schulman2017proximal} within an asymmetric actor–critic setup~\cite{pinto2017asymmetric}, with hyperparameters listed in \tabref{tab:hyperparams_rl} and an adaptive learning rate~\cite{rudin2022learning}. To mitigate the sim-to-real gap, we add standard Gaussian noise to the inputs of the actor and small disturbance forces listed in \tabref{tab:dist_forces}. In addition to the standard observations, the critic receives privileged observations consisting of noiseless quantities, friction parameters, rigid body velocities and accelerations, and the phase of the episode. Both the policy and value function are modeled using multi-layer perceptron (MLP) networks with ELU activations~\cite{clevert2016fast}, consisting of three layers with $512$ units each. Observations are normalized using a running mean, following the standard practice in PPO~\cite{schulman2017proximal}. Our simulations are performed using the GPU-accelerated Isaac Sim physics engine~\cite{NVIDIA_Isaac_Sim}, running $4096$ environment instances in parallel on a single RTX $4090$ GPU. We train our falling policy for $75$k iterations (approx. \SI{48}{\hour}).

Next, we detail additional aspects of the reward formulation.  
When resolving contacts, the physics engine can generate excessively large forces, which is why, during training only, we clip values above \SI{10}{\kilo\newton} to improve numerical stability.
To account for varying sensitivity across robot components, we assign different sensitivity weights. Previous experiments with our bipedal robot revealed that most damage occurred at the head, followed by the shoulders and elbows. To reflect this varying sensitivity, we assign the following sensitivity weights to the body parts: The pelvis and legs are weighted $1.0$, elbows $2.0$, shoulders $3.0$, and head $4.0$. In ~\secref{sec:critical_comp}, we perform an ablation of this weighting scheme.

\subsection{Experimental Details}
\label{sec:experimental_details}

\subsubsection{Data}

Our training dataset comprises $24$k target end poses, complemented by a test set of $2$k poses, generated through our sampling-based generation (\secref{sec:procedural}). Additionally, we employ a set of $10$ expressive, artist-designed poses in various orientations to explore the generality of our method. These poses were manually created by artists in Blender~\cite{blender}, respecting the joint limits and avoiding self-penetration during the process, but ignoring physical constraints. The end poses shown in \figref{fig:artistic} are visual examples of the artist-designed poses in simulation. In \secref{sec:aba_data_size}, we analyze the effect of dataset size on policy training.
\subsubsection{Metrics}
We base our evaluation on \cite{QP_Samy2017,AlmaFall_Ma2023} by comparing damage criteria across different control strategies. Unless specified otherwise, we evaluate the metrics over $32768$ trials with randomly sampled initial states and unseen target end poses from the test set. We use the following metrics throughout our experiments:
\begin{itemize}
    \item \textbf{Max Impact Force:} The maximum impact force experienced by the robot during each rollout and across body parts.
    \item \textbf{Mean Impact Force:} The maximum of the mean impact force over all body parts experienced by the robot during each rollout.
    \item \textbf{Mean Root Orientation Error (MROE):}
    The mean root orientation error is given as the geodesic distance between the global yaw axis-aligned target end pose orientation and the robot’s root orientation at the final time step of an episode.
    \item \textbf{Mean Joint Tracking Error (MJE):} The mean absolute joint tracking error over all joints at the last timestep of an episode.
\end{itemize}
\subsubsection{Robot}
We run experiments on a custom-built bipedal robot with $20$ degrees of freedom (DoF), a total mass of \SI{16.2}{\kilogram} and a height of \SI{0.84}{\meter}. Each leg has $5$ DoF with Unitree A1 actuators, and the arms and neck are equipped with Dynamixel XH540-V150-R actuators. Our policy predicts actuator positions at \SI{50}{\hertz} that are passed to proportional-derivative (PD) controllers at each joint. We estimate the robot's state by fusing information from an onboard inertial measurement unit and motion capture.

\begin{figure}[t!]
 \centering
 \includegraphics[width=\linewidth]{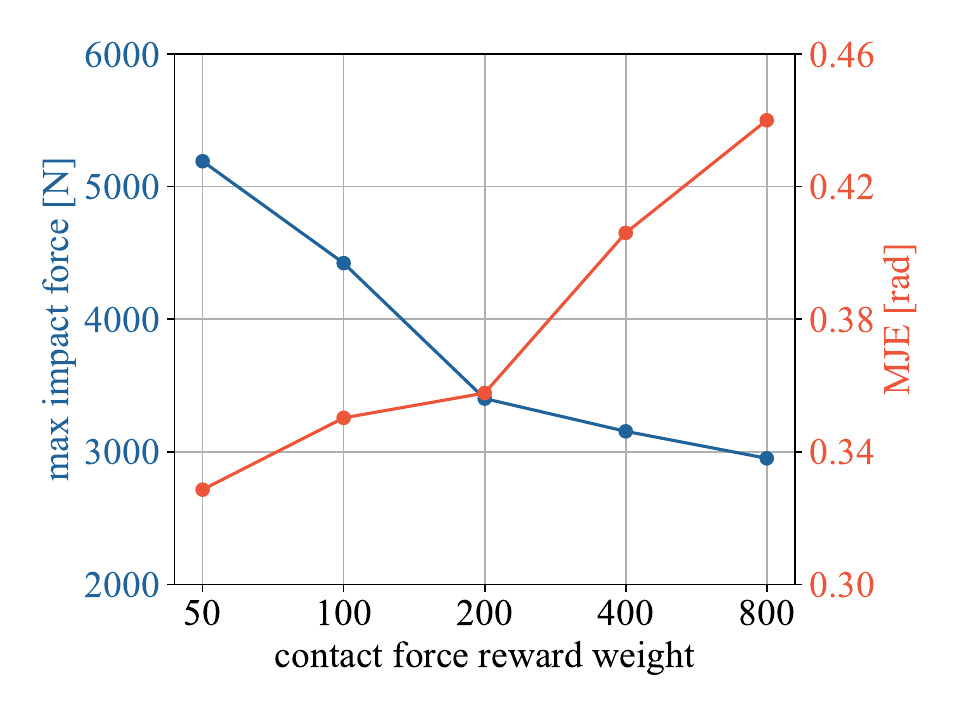}
  \caption{\textbf{Impact vs.~Tracking Ablation.} We measure the max impact force and mean joint tracking error for varying impact reward weights. Displayed are the mean values over all trials.}
 \label{fig:track_abl}
\end{figure}

\begin{table}[tb]
\begin{center}
    \caption{\textbf{Sampling-Based End Pose Ablation}. We compare two variants of our method, once trained with our sampling-based end pose generation (\emph{generated}) and once trained on artist-designed poses (\emph{artistic}). We report the mean and standard deviation of the mean joint position tracking error (MJE) and the mean root orientation error (MROE).}
    \label{tab:tracking_eval}
    \begin{tabular}{l l c c}
    \toprule
    \textbf{Test Split} & \textbf{Training Data} & \textbf{MJE} [\si{rad}] & \textbf{MROE} [\si{rad}] \\
    \midrule
    \midrule
    \multirow{2}{*}{Generated} 
        & Generated  & $0.36 \pm 0.10$ & $0.12 \pm 0.12$ \\
        & Artistic & $1.03 \pm 0.20$ & $1.05 \pm 0.58$ \\
    \midrule
    \multirow{2}{*}{Artistic} 
        & Generated & $0.30 \pm 0.09$ & $0.09 \pm 0.07$ \\
        & Artistic (seen)& $0.17 \pm 0.12$ & $0.08 \pm 0.15$ \\
    \bottomrule
    \end{tabular}
\end{center}
\end{table}

\subsection{Comparison with Standard Falling Methods}
\label{sec:comparison_failsafe}

Following the evaluations in~\cite{atkeson2015nofalls, LearningUnified_Kumar2017, AlmaFall_Ma2023}, we compare our method against standard falling strategies commonly used in practice: applying zero torque, damping the actuators with low gains ($0.1\times$nominal), and freezing them with high gains ($10\times$nominal) at their most recent setpoints. As shown in \figref{fig:impact_force}, our method substantially reduces both the max and mean impact forces compared to the baselines and exhibits much lower variance. Furthermore, the falling dynamics with our method are controlled and predictable. In contrast, freezing the joints makes the robot behave as a single rigid body, falling in the initial direction, while damping or zero-torque settings produce interactions between components, resulting in more complex and less predictable motion. Please refer to our supplemental video for visual evidence of these insights. Overall, these results highlight the benefits of our approach over common existing falling strategies.

\begin{figure}[t!]
  \centering
  \includegraphics[width=\linewidth]{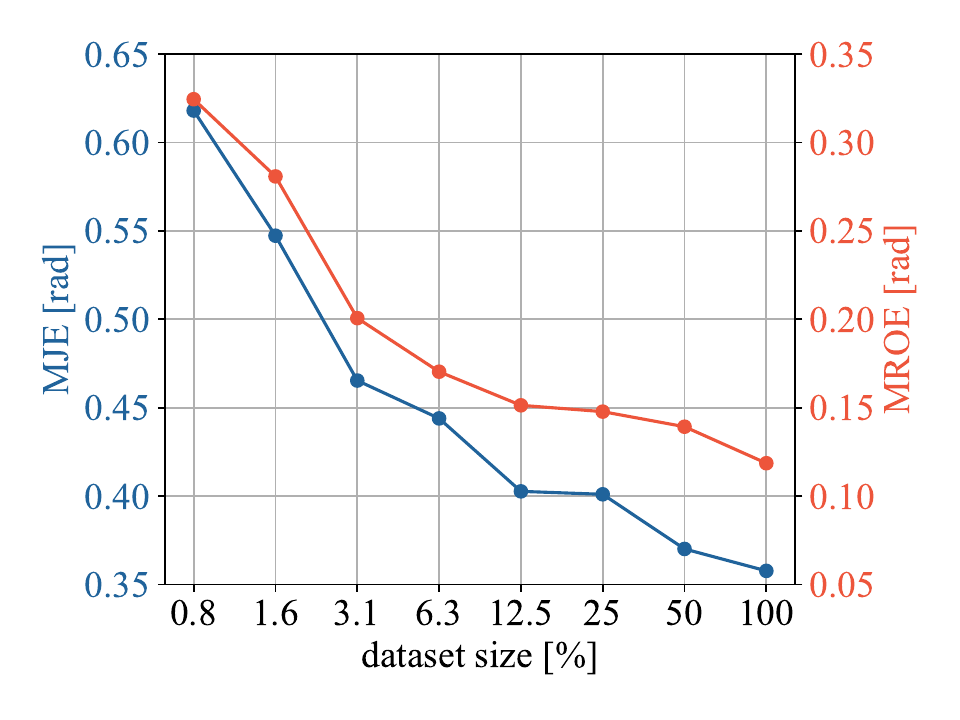}
  \caption{\textbf{Dataset Size.} Mean joint tracking error and mean root orientation error for varying dataset sizes.}
  \label{fig:data_ablation}
\end{figure}

\subsection{Ablations}
\label{sec:ablations}

\subsubsection{Impact vs. Tracking Ablation}
In this experiment, we ablate policies trained with varying weights of the contact force reward (see \tabref{tab:rewards}) and evaluate the resulting maximal impact forces and mean joint tracking error. We illustrate the results in \figref{fig:track_abl}. As expected, increasing the contact force weight reduces impact forces but increases the joint tracking error. This highlights the inherent trade-off between minimizing impact and accurately reaching the target end pose. We found that a contact force weight of $200$ provides a reasonable balance between these objectives.

\begin{figure*}[t!]
 \centering
  \begin{minipage}{\linewidth}
  \centering
  \begin{minipage}{0.24\linewidth}
   \centering
   \includegraphics[width=\linewidth]{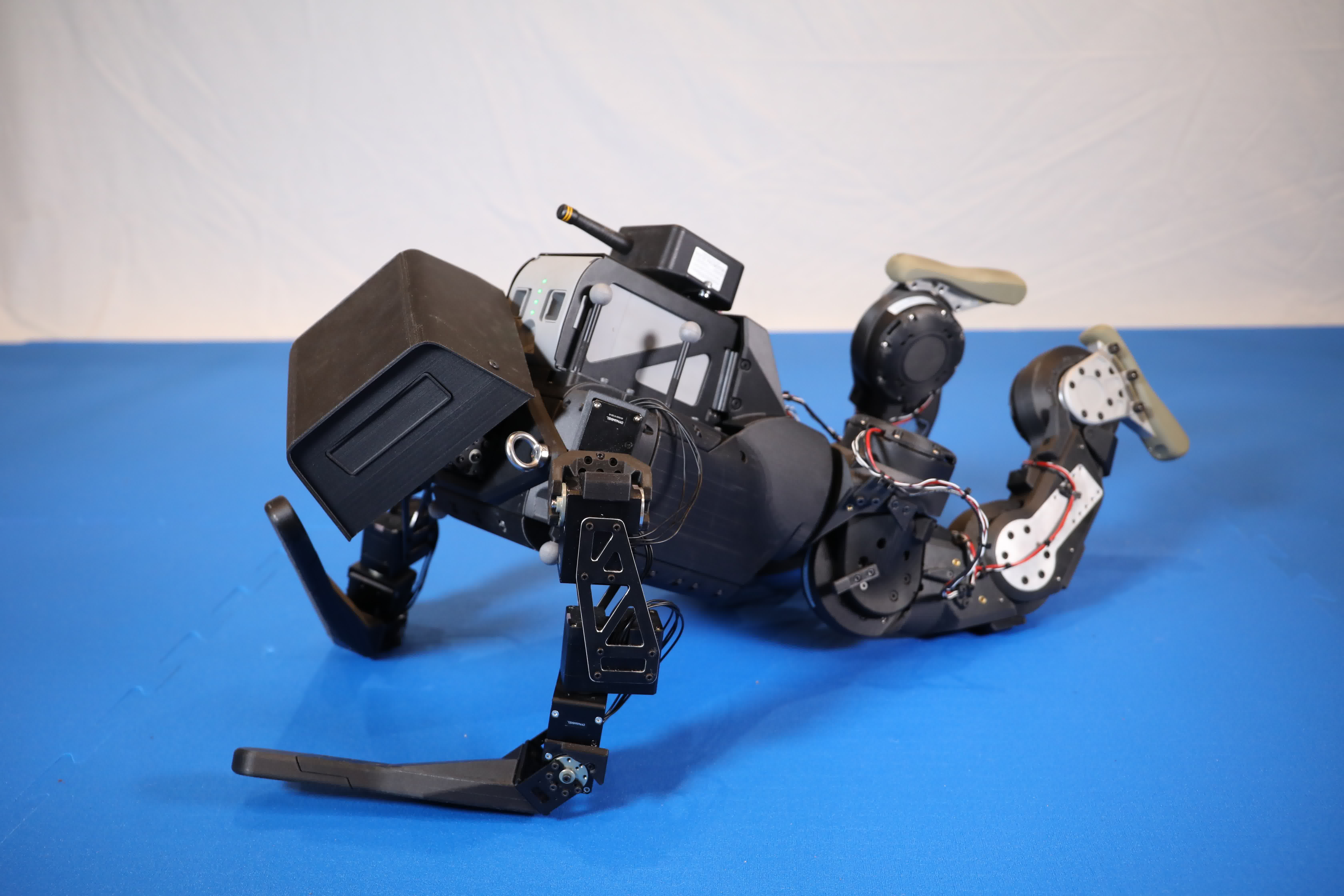}
   \\ 
  \end{minipage}
  \begin{minipage}{0.24\linewidth}
   \centering
   \includegraphics[width=\linewidth]{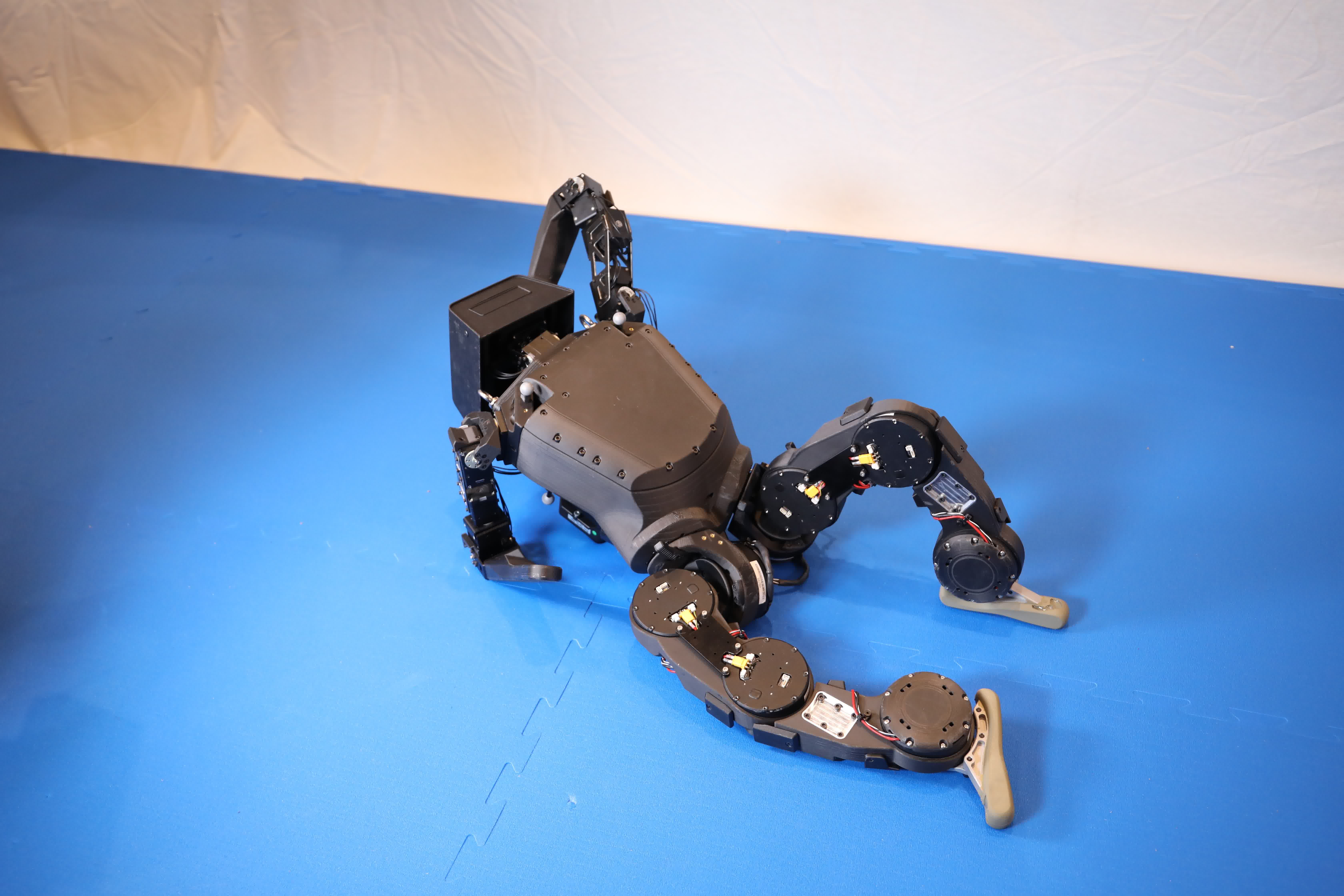}
   \\ 
  \end{minipage}
  \begin{minipage}{0.24\linewidth}
   \centering
   \includegraphics[width=\linewidth]{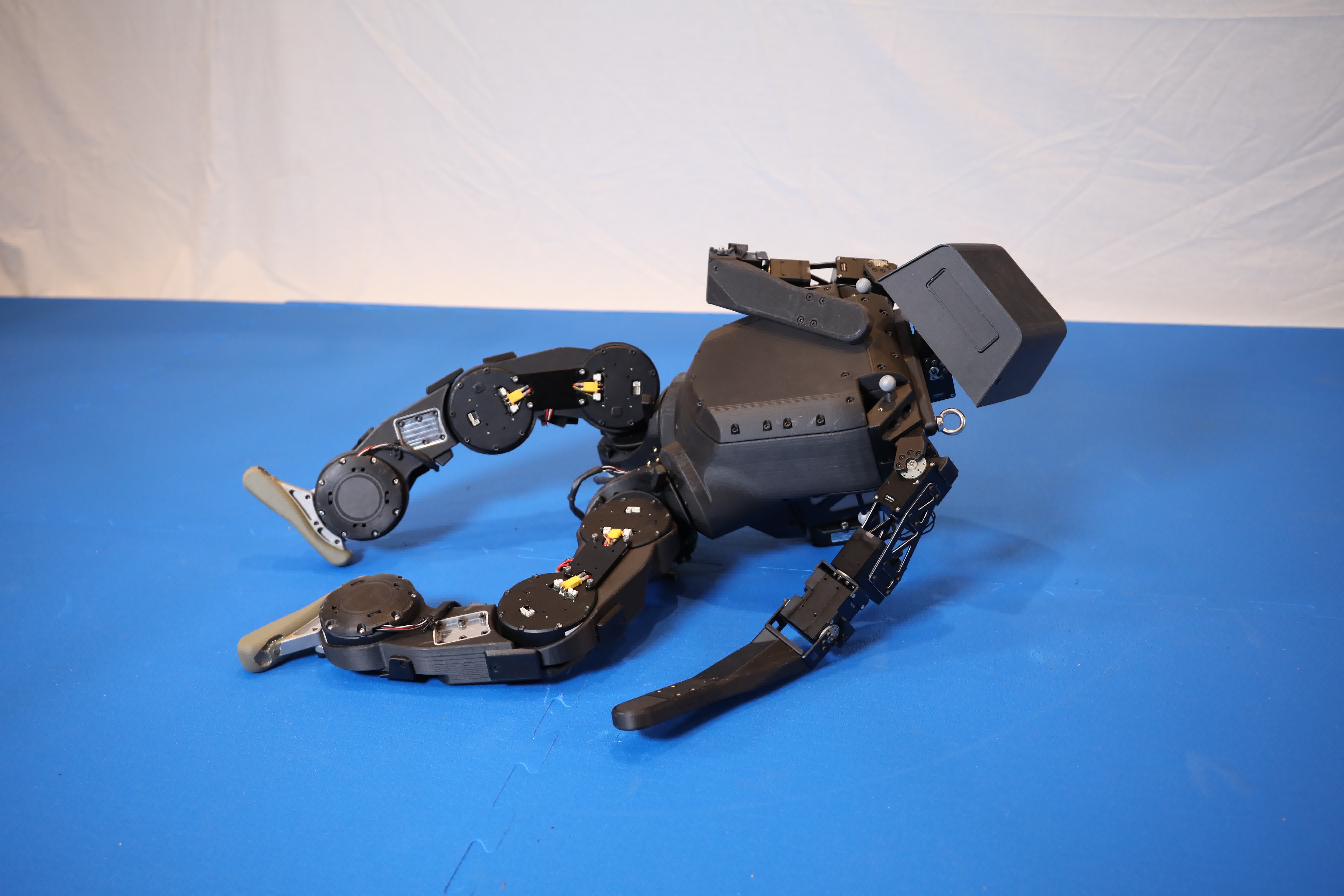}
   \\ 
  \end{minipage}
  \begin{minipage}{0.24\linewidth}
   \centering
   \includegraphics[width=\linewidth]{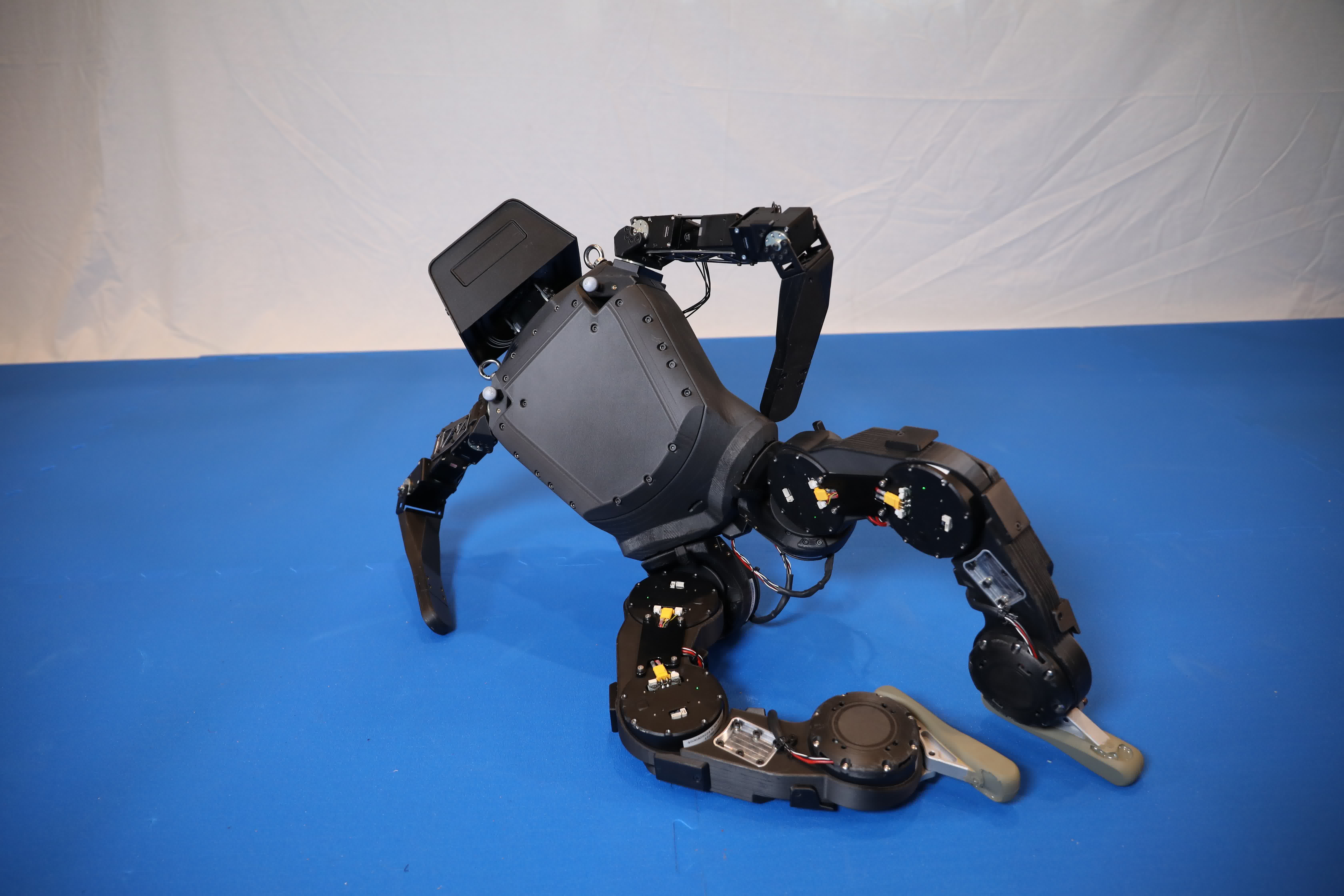}
   \\ 
  \end{minipage}
 \end{minipage}
 \caption{\textbf{Real-World Experiments.} Qualitative examples of the artist-designed end poses the robot reached after falling.}
 \label{fig:qual_real}
\end{figure*}

\subsubsection{Sampling-Based End Pose Generation}
To evaluate our sampling-based end pose generation technique (see \secref{sec:procedural}), we compare our approach trained on sampled end poses (\emph{generated}) with a variant trained solely on artist-designed end poses (\emph{artistic}). We report results in \tabref{tab:tracking_eval}. On the \emph{generated} test set, the policy trained on generated end poses outperforms the variant trained on a few artist-designed end poses significantly in both mean joint tracking error and mean root orientation error.

On the other hand, our method trained on generated end poses achieves slightly higher mean root orientation and joint tracking errors on the artist-designed test set. Note, however, that the artist-trained variant has seen these poses during training. This leads to overfitting as indicated by its high errors on the generated test dataset. In contrast, our method generalizes well to unseen poses, even when drawn from a different data distribution (\emph{artistic}).

\begin{table}[t!]
\begin{center}
    \caption{\textbf{Impact Reduction of Critical Components.} Comparison of the baseline and the policy trained with increased sensitivity weight on the battery. We report the mean joint tracking error (MJE) and mean root orientation error (MROE) as mean with standard deviation, and battery impact forces as median and 95th percentile, reflecting their highly non-symmetric statistics.
}
    \label{tab:impact_eval}
    \begin{tabular}{l c c c}
    \toprule
    \textbf{Policy} & \textbf{MJE [\si{rad}]} & \textbf{MROE [\si{rad}]} & \textbf{Median/95th \% [\si{N}]}\\
    \midrule
    \midrule
    w/o battery & $0.32 \pm 0.10$ & $0.11 \pm 0.11$ & $36.12 / 3321.75$ \\
    w/ battery  & $0.42 \pm 0.11$ & $0.16 \pm 0.14$ & $0.00  / 810.69$ \\
    \bottomrule
    \end{tabular}
\end{center}
\end{table}

\subsubsection{Dataset Size}
\label{sec:aba_data_size}
We examine how the number of generated end poses affects generalization to unseen end poses. We train multiple variants of our method, each using a progressively smaller subset of the full training dataset. We report the mean joint tracking error and mean root orientation error on our test set of unseen end poses, and illustrate the results in \figref{fig:data_ablation}. We find that the best performance is achieved with our full dataset, yielding improved joint and orientation tracking. Dataset size is most critical in low-data regimes ($1$\%-$6$\% of the total dataset), indicating that a minimum amount of data is needed for generalization. Beyond this range, additional data continues to improve performance; however, the gains become more marginal.

\subsection{Impact Reduction of Critical Components}
\label{sec:critical_comp}
Our method accounts for the varying sensitivity of different robot components. To evaluate this formulation, we split the pelvis into the main body and a rear battery pack, assigning a high sensitivity weight of $5.0$ to the battery and $1.0$ to all other components. This simulates a robot carrying sensitive hardware on its back. We compare a policy trained with these weights to a policy that has all sensitivity weights set to $1.0$. The results in \tabref{tab:impact_eval} show a significant reduction in the 95th percentile, demonstrating that worst-case impacts can be greatly reduced. A median of $0.0$ indicates that, in most falling scenarios, forces on the backpack can be fully eliminated. Thus, our method provides a general framework to balance tracking performance and impact forces on critical components. We provide qualitative results of this experiment in our supplemental video.

\subsection{Real-World Experiments}
\label{sec:sim2real}

We perform a set of qualitative real-world experiments to demonstrate the transferability of our method from simulation to the real world using the bipedal robot described in \secref{sec:experimental_details}). We select $10$ artist-designed end poses and vary the initial conditions by randomly applying external forces to the robot with a stick. We then record the resulting falling behavior, with end poses illustrated in \figref{fig:qual_real} and the entire falling motions shown in our supplemental video. Notably, we performed all of our experiments with a single robot, which remained fully functional throughout the experiments and showed no noticeable damage. This indicates that our method enables soft falling behavior that protects the robot's most sensitive part, regardless of the falling direction.

\section{DISCUSSION}
\label{sec:discussion}
Our approach shows promising results for bipedal falling, but has several limitations. Our experiments were all carried out with the same humanoid robot. While our modeling is agnostic to the robot morphology, future research could explore how well our method transfers to different humanoids or legged robots in general. 

For testing purposes, we study falling in an isolated manner and intentionally place the robot in unstable states that result in falling. A practical, real-world deployment of our approach would require a mechanism that predicts unstable states to trigger appropriate falling motions. To anticipate a fall, simple heuristics, such as detecting invalid state estimates, insufficient battery, or other safety-critical conditions, could be used. Future work could explore predicting a fall from the robot’s motion dynamics. 

In our approach, the impact weight per robot part must be defined prior to training. An exciting future avenue is the exploration of a policy that enables the adjustment of the policy's objectives at inference time, similar to multi-objective RL approaches~\cite{alegre2025amor}. This would allow users to, for example, increase the weight of impacts on components that are nearing their wear-and-tear limits.

Moreover, in our presented experiments, we pre-selected the target end poses. An interesting direction for future work is to automatically determine the most suitable falling pose based on the robot's initial state. 

Finally, while we focused on stylized and soft falling, this behavior is tightly coupled with recovery, which has been explored in recent works \cite{tao2022getup,he2025learning}. Future work could investigate how to best combine the training of falling and recovery policies, taking stylization into consideration in both policies. 

\section{CONCLUSION}

Falling remains an inevitable possibility for legged robots, and in this work, we have shown that a purpose-trained RL controller is able to both reduce the impact and severity of the fall, and also prescribe the pose in which the robot ends up. We have evaluated our method with both simulated and real-world experiments.

Falling means to temporarily relinquish control of the system. However, if the final state of the fall can be controlled, and damage can be mitigated, this also opens the potential for deliberately exploiting falls during robot operation. This could be applicable for stunt robots and slapstick performances, but could also be exploited in the future to traverse more extreme terrain.

\addtolength{\textheight}{-0cm}   





\section*{ACKNOWLEDGMENT}

We sincerely thank Violaine Fayolle and Doriano van Essen for designing the artistic poses used in this project.


\bibliography{bibliography.bib}
\bibliographystyle{IEEEtran}

\end{document}